\documentclass[sigconf]{acmart}
\makeatletter
\def\@ACM@checkaffil{
    \if@ACM@instpresent\else
    \ClassWarningNoLine{\@classname}{No institution present for an affiliation}%
    \fi
    \if@ACM@citypresent\else
    \ClassWarningNoLine{\@classname}{No city present for an affiliation}%
    \fi
    \if@ACM@countrypresent\else
        \ClassWarningNoLine{\@classname}{No country present for an affiliation}%
    \fi
}
\makeatother

\AtBeginDocument{%
  }

\setcopyright{acmcopyright}
\copyrightyear{2023}
\acmYear{2023}
\acmDOI{XXXXXXX.XXXXXXX}

\acmConference[MM'23]{Make sure to enter the correct
  conference title from your rights confirmation emai}{Oct. 29--Nov. 3, 2023}{Ottawa, Canada}
\acmPrice{15.00}
\acmISBN{978-1-4503-XXXX-X/18/06}

\renewcommand\footnotetextcopyrightpermission[1]{}
\acmSubmissionID{429}

\usepackage{multirow}
\usepackage{multicol}
\usepackage{subcaption}

\usepackage[misc]{ifsym}

\newcommand{\ie}{\textit{i}.\textit{e}., }
\newcommand{\eg}{\textit{e}.\textit{g}., }

\begin{document}

\title{Ada-DQA: Adaptive Diverse Quality-aware Feature Acquisition for Video Quality Assessment}

\author{Hongbo Liu$^{\dag}$}
\email{liuhbleon@gmail.com}
\affiliation{
  \institution{Tsinghua University}
}

\author{Mingda Wu$^{\dag}$}
\email{wumingda@kuaishou.com}
\affiliation{
  \institution{Kuaishou Technology}
}

\author{Kun Yuan$^{\dag~\textrm{\Letter}}$}
\email{yuankun03@kuaishou.com}
\affiliation{
  \institution{Kuaishou Technology}
}

\author{Ming Sun}
\email{sunming03@kuaishou.com}
\affiliation{
  \institution{Kuaishou Technology}
}

\author{Yansong Tang}
\email{tang.yansong@sz.tsinghua.edu.cn}
\affiliation{
  \institution{Tsinghua University}
}

\author{Chuanchuan Zheng}
\email{zhengchuanchuan@kuaishou.com}
\affiliation{
  \institution{Kuaishou Technology}
}

\author{Xing Wen}
\email{wenxing@kuaishou.com}
\affiliation{
  \institution{Kuaishou Technology}
}

\author{Xiu Li$^{~\textrm{\Letter}}$}
\email{li.xiu@sz.tsinghua.edu.cn}
\affiliation{
  \institution{Tsinghua University}
}

\begin{abstract}

Video quality assessment (VQA) has attracted growing attention in recent years. While the great expense of annotating large-scale VQA datasets has become the main obstacle for current deep-learning methods. To surmount the constraint of insufficient training data, in this paper, we first consider the complete range of video distribution diversity (\ie content, distortion, motion) and employ diverse pretrained models (\eg architecture, pretext task, pre-training dataset) to benefit quality representation. An Adaptive Diverse Quality-aware feature Acquisition (Ada-DQA) framework is proposed to capture desired quality-related features generated by these frozen pretrained models.  By leveraging the Quality-aware Acquisition Module (QAM), the framework is able to extract more essential and relevant features to represent quality. Finally, the learned quality representation is utilized as supplementary supervisory information, along with the supervision of the labeled quality score, to guide the training of a relatively lightweight VQA model in a knowledge distillation manner, which largely reduces the computational cost during inference. Experimental results on three mainstream no-reference VQA benchmarks clearly show the superior performance of Ada-DQA in comparison with current state-of-the-art approaches without using extra training data of VQA.


\end{abstract}

\keywords{video quality assessment, diverse pretrained model, quality-aware representation, sparsity constraint, knowledge distillation}



\begin{teaserfigure}
\begin{center}
\includegraphics[width=\textwidth]{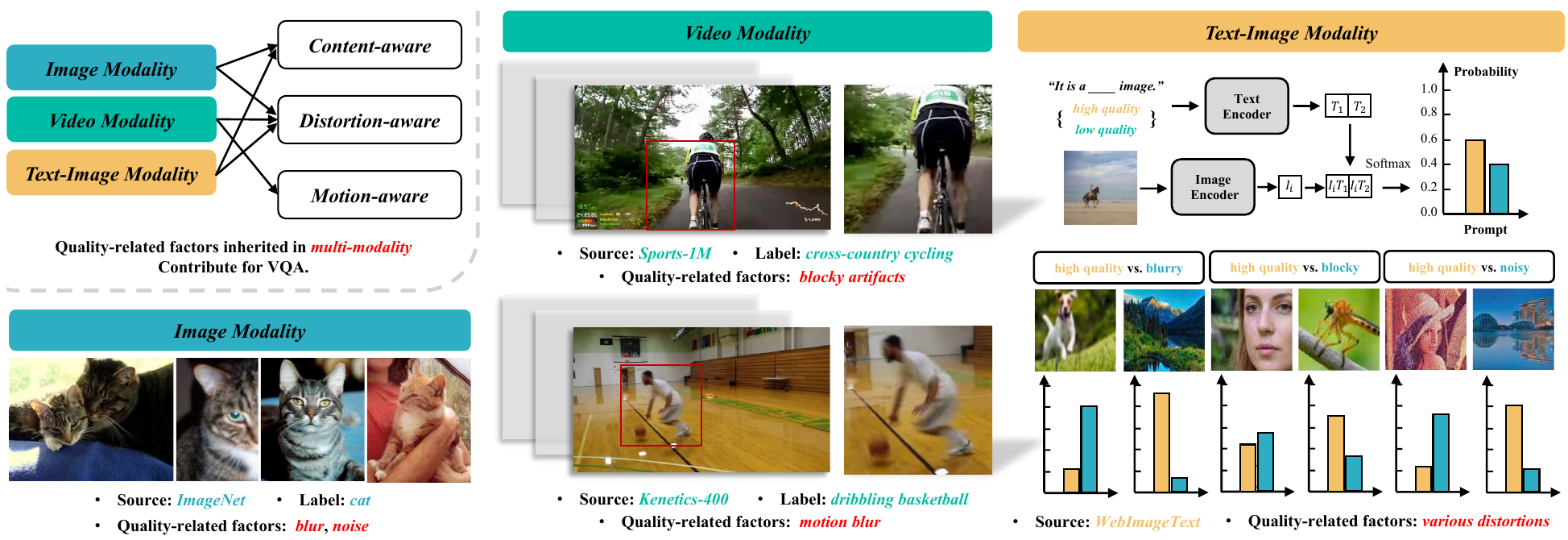}
\end{center}
\caption{
Quality-related factors (\ie content, distortion, and motion) are encompassed within the information of different modalities. For the image modality (\eg ImageNet \cite{DBLP:conf/cvpr/DengDSLL009/imagenet}), information pertaining to content and distortion perception is included. For the video modality (\eg Sports-1M \cite{KarpathyCVPR14/sports-1m}, Kinetics-400 \cite{DBLP:journals/corr/KayCSZHVVGBNSZ17/k400}), compression distortions or motion blur are taken into account. For the cross-modality (\eg WebImageText \cite{DBLP:conf/icml/RadfordKHRGASAM21/clip}, emotional descriptions associated with visual quality may be contained.
}
\label{fig:intro}
\end{teaserfigure}

\maketitle

\def\thefootnote{\dag}\footnotetext{Equal contribution. $^{\textrm{\Letter}}$ Corresponding authors. This work was done when Hongbo Liu was a student intern at Kuaishou Technology.}

\section{Introduction} \label{sec:intro}

With the explosive growth of video content-based social media, there has been a tremendous amount of videos produced and shared \cite{NBS_2021/cisco}. To guarantee optimal video quality and ensure users' quality of experience (QoE), VQA plays a crucial role in guiding image processing and video coding systems \cite{DBLP:conf/cvpr/ChadhaA21,DBLP:conf/mm/XuLZZW021/stdm}. Benefit from the thrive of Deep Neural Networks (DNN), DNN-based VQA methods \cite{DBLP:conf/mm/LiJJ19/vsfa,DBLP:journals/ijcv/LiJJ21,Ying_2021_CVPR/patch-vq,DBLP:journals/access/Gotz-HahnHLS21/konvid150k,DBLP:conf/icip/YouK19,DBLP:conf/mm/KorhonenSY20} have shown great results on in-the-wild VQA benchmarks \cite{DBLP:journals/tip/SinnoB19/live-vqc, DBLP:conf/qomex/HosuHJLMSLS17/konvid, DBLP:conf/mmsp/WangIA19/youtube-ugc, Ying_2021_CVPR/patch-vq}.



DNN-based VQA methods are hindered by the limited scale of existing VQA datasets. As demonstrated in Tab.~\ref{tab:annotation}, public VQA datasets are significantly smaller in size when compared to video classification datasets \cite{KarpathyCVPR14/sports-1m,DBLP:journals/corr/KayCSZHVVGBNSZ17/k400}. The primary reason for this discrepancy is the prevalent use of crowdsourcing to annotate subjective video quality \cite{DBLP:journals/network/ChenCWCL10,DBLP:journals/tmm/HossfeldKHGHDT14,DBLP:conf/qomex/ShahidSPBWTG14,DBLP:journals/comsur/ChenWZ15}, which is a time-consuming yet indispensable process in eliminating randomness and enhancing consistency. For example, the KoNViD-1k dataset \cite{DBLP:conf/qomex/HosuHJLMSLS17/konvid} requires an average of 114 subjective scores to produce a valid label. Recently, there have been attempts to scale up VQA datasets \cite{Ying_2021_CVPR/patch-vq,DBLP:journals/access/Gotz-HahnHLS21/konvid150k}. \textbf{However, these efforts often sacrifice the annotation quality, as they reduce the average number of annotations per video in order to improve efficiency.} For instance, an experiment conducted on the KoNViD-150k dataset \cite{DBLP:journals/access/Gotz-HahnHLS21/konvid150k} revealed that the correlation coefficient of SRCC between randomly sampled five annotations is 0.8, while it is 0.9 for fifty annotations. Thus, in this paper, we primarily focus on enhancing the performance of VQA under the circumstance of limited availability of high-quality annotated data.


\begin{table}[t]
    \centering
    \caption{Comparisons of public VQA and video classification datasets. Note that obtaining reliable quality annotations requires massive subjective annotations per video, which limits the size of VQA datasets.}
    \begin{tabular}{c|c|c|c}
    \toprule
    Dataset & Task & Size & Annotations \\
    \midrule
    KoNViD-1k \cite{DBLP:conf/qomex/HosuHJLMSLS17/konvid} & VQA & 1,200 & 114 \\
    LIVE-VQC  \cite{DBLP:journals/tip/SinnoB19/live-vqc} & VQA & 585 & 240 \\
    YouTube-UGC \cite{DBLP:conf/mmsp/WangIA19/youtube-ugc} & VQA & 1,380 & 123 \\
    \midrule 
    LSVQ \cite{Ying_2021_CVPR/patch-vq} & VQA & 39,075 & 35 \\
    KoNViD-150k \cite{DBLP:journals/access/Gotz-HahnHLS21/konvid150k} & VQA & 153,841 & 5 \\
    \midrule
    Sports-1M \cite{KarpathyCVPR14/sports-1m} & classification & 1,133,158 & - (\textit{auto.}) \\
    Kinetics-400 \cite{DBLP:journals/corr/KayCSZHVVGBNSZ17/k400} & classification & 306,245 & 3-5 \\
    \bottomrule
    \end{tabular}
    \label{tab:annotation}
\end{table}

To surmount the constraint of insufficient training data, certain works \cite{DBLP:conf/mm/LiJJ19/vsfa,DBLP:journals/ijcv/LiJJ21,DBLP:journals/spic/HouZHL20/pretrain-transfer,DBLP:journals/sivp/VargaS19/pretrain-lstm} attempt to utilize DNNs that have been pre-trained on other extensive datasets, for the purpose of fine-tuning. The majority of these works employ ImageNet \cite{DBLP:conf/cvpr/DengDSLL009/imagenet}, which comprises a vast array of object categories, as a pre-training dataset. By introducing content-aware knowledge which meets the diverse distribution of in-the-wild videos, a considerable enhancement in performance can be obtained in downstream VQA tasks. However, videos captured in-the-wild often encounter unavoidable distortions resulting from extreme shooting conditions, compression, transmission, or other unprofessional operations performed by users \cite{DBLP:conf/mm/XuLZZW021/stdm}. In such scenarios, relying solely on content-aware information is insufficient to ensure quality representation. Furthermore, some other studies \cite{DBLP:conf/cvpr/WangKTYBAMY21/rich-feats, DBLP:journals/corr/abs-2108-08505/bvqa} seek to leverage motion-aware and compression-aware features obtained from models that are pretrained using spatiotemporal information, simply by concatenating or averaging the features. As shown in Fig.~\ref{fig:intro}, \textbf{quality-related factors are encompassed within the information of different modalities, and the contribution of each modality may vary depending on the input}. Nevertheless, most existing methods do not consider the complete range of video distribution diversity or employ simplistic methods to integrate various types of features, which restricts further improvements in VQA.



To address the aforementioned limitations, in this paper, we aim to employ diverse in-the-wild pretrained models to enhance VQA performance from various aspects that may affect video quality. In the case of the image modality, owing to the presence of content labels and a vast number of images closely resembling the actual distribution, quality-related information pertaining to content and distortion perception is included. As for the video modality, the data distribution contains quality-related information such as compression distortions (\eg blocky artifacts) or motion blur. For the cross-modality of image-text, we use a trained CLIP model \cite{DBLP:conf/icml/RadfordKHRGASAM21/clip}, which efficiently learns visual concepts from natural language supervision, for analysis. By devising appropriate templates, CLIP exhibits remarkable predictive ability without access to VQA annotations. We surmise that during the training of CLIP, certain texts may contain emotional descriptions associated with the quality. \textbf{These phenomena attest to the practicality of employing in-the-wild models that have been pretrained on diverse modalities of data}.

Based on the above observations, in this paper, we introduce a new \textbf{Adaptive Diverse Quality-aware feature Acquisition (Ada-DQA)} framework for VQA tasks. Initially, we establish a pool of pre-trained models exhibiting a wide range of diversity, taking into account various aspects of their architectures, pretrained pretext tasks, and pretrained datasets. Subsequently, to dynamically capture the desired quality-related features on a per-sample basis during training, we present a \textbf{Quality-aware Acquisition Module (QAM)} that outputs gating weights of features generated by different pretrained models for aggregation. We also impose an additional sparsity constraint on the distribution of the gating weights, encouraging attention to be focused on more crucial and pertinent features for quality representation. Lastly, the learned quality representation is employed as auxiliary supervisory information, in conjunction with the supervision of the labeled quality score, to facilitate the training of a comparatively lightweight VQA model using knowledge distillation. The effectiveness of our method is evaluated through extensive experiments on three mainstream high-quality annotated NR-VQA datasets. Adapt-DQA models achieve 0.8651, 0.8591, and 0.8729 of SRCC on KoNViD-1k, LIVE-VQC, and YouTube-UGC, improving the \textit{state-of-the-art} results for these datasets by absolute margins of 0.6\%, 1.79\%, and 3.88\%, respectively. 
Our \textbf{contributions} are as follows:
\begin{itemize}
\item To the best of our knowledge, this is the first study to comprehensively investigate the relationships between pretrained models and video quality. We construct a diverse pool that encompasses a broad spectrum of quality-related factors.
\item We propose the Ada-DQA to leverage these pretrained models for VQA, where the QAM is proposed to capture quality-related features adaptively. Additionally, a sparsity constraint is also attached for the most crucial and relevant features.
\item We evaluate Ada-DQA on three mainstream NR-VQA benchmarks, surpassing current \textit{state-of-the-art} methods without using extra training data of QA. Sufficient ablation studies validated the effectiveness of each component.
\end{itemize}

\section{Related work}

According to the availability of reference videos, VQA methods can be classified into full-reference (FR), reduced-reference (RR), and no-reference (NR) \cite{DBLP:journals/corr/abs-2108-08505/bvqa} ones. As reference videos are always hard to obtain, NR-VQA becomes more practical in the real-world VQA scenario, which is investigated in this paper. According to the difference in construction schema, VQA methods can be classified into traditional hand-crafted and learning-based ones.

\begin{figure*}[t]
\begin{center}
\includegraphics[width=\textwidth]{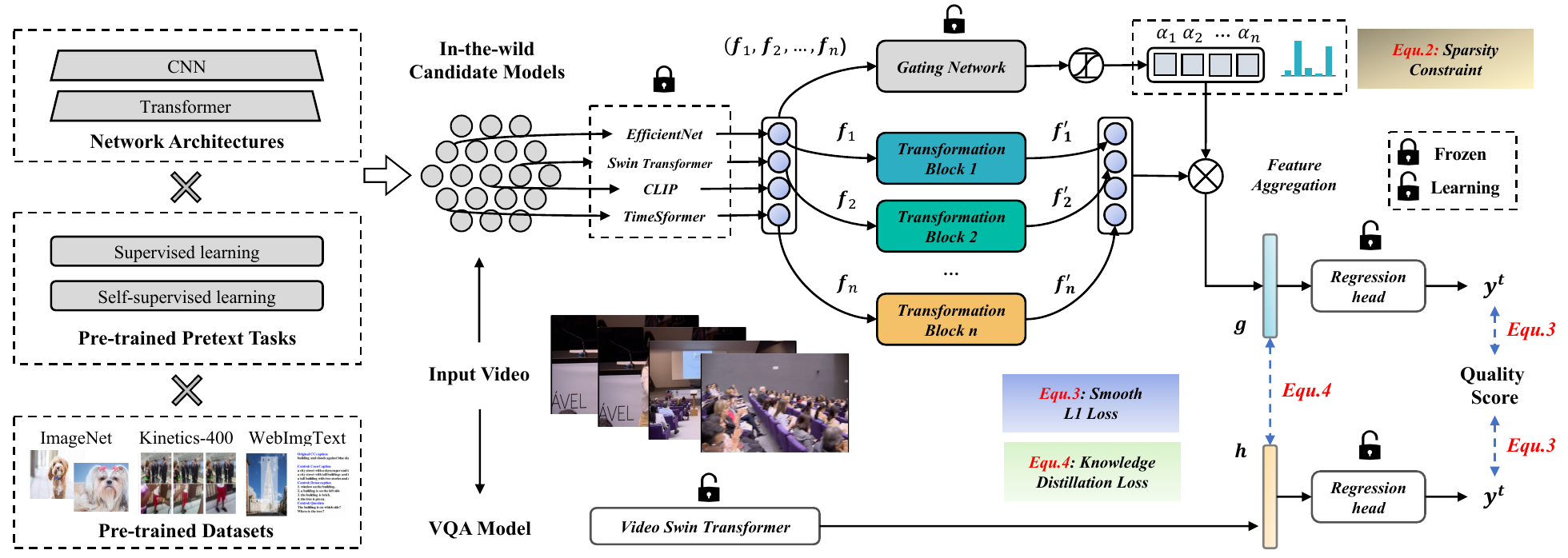}
\end{center}
\caption{
Overview of our proposed Ada-DQA framework. First, in-the-wild pretrained models are selected as candidates according to diverse aspects. Second, features generated by these frozen pretrained models are aggregated per sample using the QAM adaptively. This approach allows acquiring of quality-related representations. Third, during training, the integrated feature is utilized as supplementary supervision, along with the labeled quality score, to guide the training of a lightweight VQA model. During inference, only the optimized VQA model is used, reducing the computational cost largely.
}
\label{fig:DQA}
\end{figure*}

\subsection{Classical VQA Approaches}

Classic VQA methods \cite{DBLP:journals/tip/SaadBC12,DBLP:journals/tip/MittalSB16/viideo,DBLP:journals/tip/Korhonen19,DBLP:journals/tip/TuWBAB21/videval,DBLP:conf/pcs/TuCWBAB21,DBLP:conf/mm/LiaoXWCSYL22} rely on handcrafted features to evaluate video quality. With an underlying assumption that the perceptual quality can be measured by the disturbance of natural scene statics (NSS) \cite{DBLP:conf/accv/YanZF16}, these work attempts at designing handcrafted features with richer representation for VQA. The work \cite{DBLP:journals/tip/SaadBC14/blind-vqa} based on the 2D discrete-time transform (DCT) features of video frame-difference statistics, and motion information is further introduced to level up the representation capacity. TLVQM \cite{DBLP:journals/tip/Korhonen19} utilizes a combination of spatial high-complexity and temporal low-complexity handcraft features. Whereas, handcrafted features are gradually replaced by the DNN-based features, due to their sensitivity to distortion types and the superiority DNN features demonstrated in various computer vision tasks.

\subsection{DNN-based VQA Methods}

Recently, CNN-based methods \cite{DBLP:conf/mm/LiJJ19/vsfa,DBLP:journals/tip/TuWBAB21/videval,DBLP:journals/pr/GuMXP19,DBLP:conf/aaai/GuMDXP19} and Transformer-based methods \cite{DBLP:journals/corr/abs-2108-05997/musiq,DBLP:journals/corr/abs-2108-09635/star-vqa,Zhao_2023_CVPR_Zoom,DBLP:conf/eccv/WuCHLWSYL22} have taken the lead in the QA domain. However, due to the data-driven characteristics of deep learning, most of the current VQA models suffer from the lack of sufficient high-quality-labeled datasets. There are some attempts to relieve this insufficient data challenge, either from patch-level/frame-level augmentation \cite{DBLP:conf/mm/LiJJ19/vsfa,DBLP:conf/mm/You21/lsct} or fine-tuning from other large computer vision models pretrained on large general knowledge-based datasets \cite{DBLP:journals/ijcv/LiJJ21,DBLP:journals/spic/HouZHL20/pretrain-transfer,DBLP:journals/sivp/VargaS19/pretrain-lstm}. VSFA \cite{DBLP:conf/mm/LiJJ19/vsfa} extracts frame-wise features with ResNet and uses a gated-recurrent unit to model temporal information. LSCT \cite{DBLP:conf/mm/You21/lsct} adopts a Transformer to predict video quality based on the frame features extracted by an IQA model. But frame-level augmentation dismissed the effect brought by temporal concealment, which is widely noticed nowadays.

Most fine-tuning work \cite{DBLP:journals/ijcv/LiJJ21,DBLP:journals/spic/HouZHL20/pretrain-transfer,DBLP:journals/sivp/VargaS19/pretrain-lstm} utilizes models pretrained on classification datasets, where learned information mainly covers content-awareness and is not tailor for the task of VQA. 
Several work \cite{DBLP:conf/cvpr/WangKTYBAMY21/rich-feats, DBLP:journals/corr/abs-2108-08505/bvqa,Zhao_2023_CVPR} has noticed the insufficiency of content-aware information. 
CoINVQ \cite{DBLP:conf/cvpr/WangKTYBAMY21/rich-feats} leverage distortion-aware and compression-aware representation besides the content-aware representation. Nevertheless, the distorted information is learned from synthetic datasets and the generalized ability to in-the-wild distorted data needs to be verified. BVQA \cite{DBLP:journals/corr/abs-2108-08505/bvqa} introduces motion-aware information learned from the action classification dataset. But they dismiss the fact that distortion awareness is crucial to VQA.
What's more, these works either utilize a temporal-sampling and concatenating strategy to aggregate features or employ temporal average pooling for feature fusion. The final features are not acquired in an adaptive and flexible manner, which prohibits the diverse feature representations from unleashing full potential. More recent work focus on building spatiotemporal relation.
StarVQA \cite{DBLP:journals/corr/abs-2108-09635/star-vqa} builds a Transformer by using divided space-time attention. DisCoVQA \cite{DBLP:journals/corr/abs-2206-09853} design a transformer-based Spatial-Temporal Distortion Extraction module to tackle temporal quality attention. FastVQA \cite{DBLP:conf/eccv/WuCHLWSYL22} attempts to assess local quality by sampling patches at their raw resolution and covers global quality with contextual relations.

\section{Method}\label{sec:method}

To surmount the constraint of limited labeled data availability and to obtain the quality-related features inherent in diverse modalities, we introduce the Ada-DQA framework for VQA tasks. In Sec.\ref{sec:framework}, we provide an overview of the framework. In Sec.\ref{sec:pretrained}, we explicate the construction of pretrained models from various aspects. In Sec.\ref{sec:moe}, we elucidate the process of acquiring quality representation using the proposed Quality-aware Acquisition Module (QAM). Finally, in Sec.\ref{sec:objective}, we present the optimization objective during training based on knowledge distillation and regression loss.

\subsection{Ada-DQA Framework}\label{sec:framework}

As shown in Fig.~\ref{fig:DQA}, the framework of Ada-DQA can be divided into three components. \textbf{First}, $N$ pretrained models, which act as feature extractors, are selected as candidates from the wild. Given an input video $\mathcal{V}$, features are generated by these pretrained models, whose weights are frozen. This significantly reduces the training cost of multiple heavy pretrained models. According to the training paradigm of pretrained models, these features may contain quality-related information (\eg, content, distortions, and motion). However, since factors that may affect quality vary in different videos, the correlation between the quality of different videos and these features is also different. \textbf{Second}, to adaptively capture desired quality-related features sample-by-sample during training, the proposed QAM is used to raise dynamic weights for feature aggregation. An extra sparsity constraint is attached to the distribution of these gating weights, promoting attention to more critical and relevant features for quality representation. Then the video quality feature can be obtained by a weighted summation. \textbf{Finally}, the learned quality representation is utilized as supplementary supervisory information, along with the supervision of the labeled quality score, to guide the training of a relatively lightweight VQA model in a knowledge distillation manner. \textbf{During inference, only the optimized VQA model is used}, reducing the computational cost largely. More details will be provided below.

\subsection{Quality-related Pretrained Models}\label{sec:pretrained}

\begin{figure}[t]
\begin{center}
   \includegraphics[width=\linewidth]{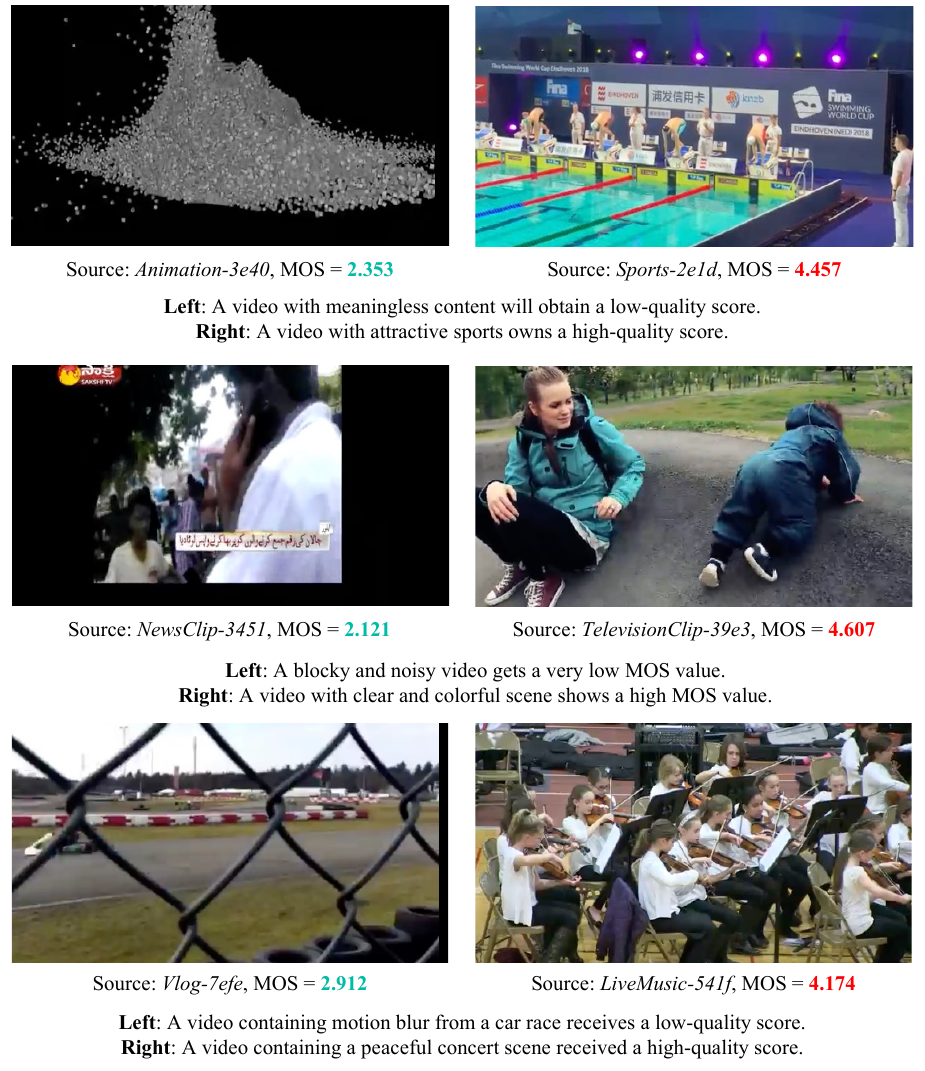}
\end{center}
\caption{Videos sampled from the YouTube-UGC dataset \cite{DBLP:conf/mmsp/WangIA19/youtube-ugc} and their corresponding labeled MOS, ranging from 1.0 to 5.0. It can be seen that video quality is affected by various aspects, including content, distortions, and motion.}
\label{fig:demo}
\end{figure}

Inspired by the current success of the ``pretraining and fine-tuning'' paradigm in deep learning \cite{DBLP:conf/cvpr/He0WXG20,DBLP:conf/icml/ChenK0H20,DBLP:conf/icml/RadfordKHRGASAM21/clip}, we aim to utilize in-the-wild pretrained models to benefit VQA from diverse aspects of the video, in order to enhance better understanding of video quality and enable personalized treatments to improve it. We contemplate the choice of pretrained models through the lens of multiple factors, as shown in Fig.~\ref{fig:demo}, that may impact the quality of videos as follows:
\begin{itemize}
    \item \textbf{Content}. Human judgments of visual quality are content-dependent according to previous studies \cite{DBLP:conf/mm/LiJJ19/vsfa,DBLP:conf/cvpr/WangKTYBAMY21/rich-feats}. When a video is visually appealing, engaging, and relevant to the viewer's interests (\eg cute puppy and beautiful scenery), it can capture their attention and make them more receptive to the video's content. In contrast, if a video is dull, uninteresting, or irrelevant (\eg black screen and messy corners), viewers are more likely to rate low quality. Introducing models pretrained on the task of object recognition (\eg EfficientNet, Swin Transformer) may benefit VQA. 
    \item \textbf{Distortion}. In addition to content, distortions introduced during the phase of video capturing and compression also determine the video quality \cite{DBLP:journals/tip/TuWBAB21}. Thus a pretrained model that has been trained on a dataset of images or videos (\eg ImageNet, Kinetics-400) with compression artifacts will have learned to identify the specific patterns and features that are associated with compression artifacts, such as blockiness, blurriness, or pixelation. 
    \item \textbf{Motion}. Unlike the image scenario, motion blur can significantly affect the quality of videos \cite{DBLP:journals/spic/WangLB04,DBLP:journals/tbc/ChikkerurSRK11}. It occurs when there is rapid motion, and the camera or objects in the scene are moving too quickly for the camera's shutter to capture. A pretrained action recognition model (\eg SlowFast, TimeSformer) may detect specific actions or movements such as running, jumping, or throwing. These can be useful for analyzing the amount of motion or by looking for specific visual cues that are associated with motion blur, such as streaking around the edges of moving objects.
\end{itemize}

\begin{figure*}[t]
\begin{center}
   \includegraphics[width=0.9\linewidth]{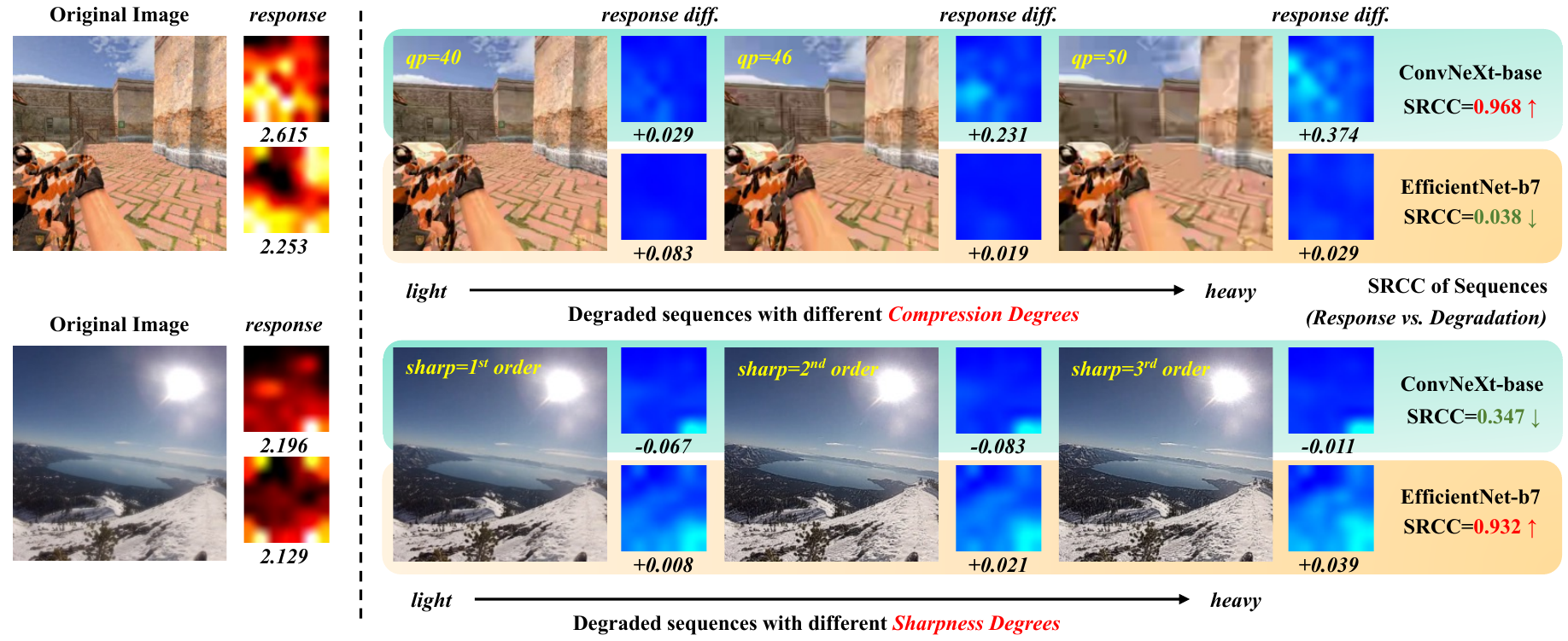}
\end{center}
\caption{Responses of different pretrained models to synthetic sequences generated by distortions (\ie compression, sharpness). The correlation of SRCC is computed according to distortion degrees. It is evident that pretrained models may detect certain types of distortions, but their ability to perceive distortion varies across models.}
\label{fig:architecture}
\end{figure*}

However, it is important to note that \textbf{individual pretrained models may not be able to identify all types of quality-related factors, or may not be as accurate in identifying certain types}. Some evidences are given in Fig.~\ref{fig:architecture}. When encountering different types of distortions, there will be obvious differences in the perception ability of the pretrained model. In detail, ConvNext-base (SRCC=0.968) outperforms EfficientNet-b7 (SRCC=0.038) when facing compression. When it comes to sharpness, EfficientNet-b7 performs better. Therefore, it is important to use a diverse set of models and combine their results to get a more robust assessment. Thus, we propose to construct a pool with a large diversity of candidate models, considering the following aspects:
\begin{itemize}
    \item \textbf{Architecture.} The efficacy of a network architecture (\eg CNN, Transformer) hinges upon its capacity to assimilate and convey information. A well-crafted architecture can discern finer details and patterns in the input video, while also influencing the manner in which spatial and temporal information, containing quality-related features, is processed.
    \item \textbf{Pretrained pretext task.} The type of supervision in the pretext task has an impact on the ability of the pretrained model to VQA tasks. When the distribution of data is similar, a supervised pretext task may lead to superior performance. Conversely, self-supervised pretext tasks, where the model is trained on unlabeled data, may facilitate better generalization when confronting unfamiliar VQA domains.
    \item \textbf{Pretrained dataset.} Pretrained datasets on a large scale can be advantageous to VQA by providing diverse content, distortion, and motion-related data. A desired pretrained dataset should include a wide range of categories that closely resemble real-world scenarios, as well as other multi-modal information that can aid in describing video quality. For instance, the WebImageText \cite{DBLP:conf/icml/RadfordKHRGASAM21/clip} dataset, which combines text and images, can be helpful in this regard.
\end{itemize}


Based on the above considerations, in this paper, we select several pretrained models that obtain top performance in their original fields, including (1) EfficientNet-b7 \cite{DBLP:conf/icml/TanL19/efficientnet} trained on ImageNet-1k \cite{DBLP:conf/cvpr/DengDSLL009/imagenet}, (2) ir-CSN-152 \cite{DBLP:conf/iccv/TranWFT19/csn} trained on Sports-1M \cite{KarpathyCVPR14/sports-1m}, (3) CLIP trained on WebImageText \cite{DBLP:conf/icml/RadfordKHRGASAM21/clip}, (4) Swin Transformer Base \cite{Liu_2021_ICCV_swin} trained on ImageNet-21k \cite{DBLP:conf/cvpr/DengDSLL009/imagenet}, (5) TimeSformer \cite{DBLP:conf/icml/BertasiusWT21/timesformer} trained on Kinetics-400 \cite{DBLP:journals/corr/KayCSZHVVGBNSZ17/k400}, (6) Video Swin Transformer Base \cite{DBLP:journals/corr/abs-2106-13230/video-swin} trained on Kinetics-600 \cite{DBLP:journals/corr/abs-1808-01340/k600}, and (7) SlowFast \cite{DBLP:conf/iccv/Feichtenhofer0M19/slowfast} trained on Kinetics-400.

\subsection{Quality-aware Acquisition Module}\label{sec:moe}

As the distribution of content and distortions in videos can be quite complex, a static combination of pretrained models may not always yield optimal performance. In order to adaptively capture the diversity and complementary information from different pretrained models, we propose a Quality-aware Acquisition Module (QAM). It takes extracted features from various pretrained models as input and produces a consolidated feature as output for the ultimate representation of quality. The computational process can be partitioned into two main parts. \textbf{The first part} is responsible for transforming the extracted features initially into a uniform feature dimension to enable subsequent aggregation. Structurally, this transformation block comprises two fully-connected layers followed by a normalization layer and a GELU activation layer. \textbf{The second part} generates gating weights $\boldsymbol \alpha$ to control the aggregation process. The gating network takes the concatenated feature vector as input and outputs a set of gating weights that represent the relative contribution of each pretrained model to the final quality representation. Structurally, this gating network is stacked using a fully-connected layer and a sigmoid layer. Then the quality representation $\mathbf{g}$ can be obtained by a weighted sum according to the gating weights. Given the extracted features by different pretrained models $\{\mathbf{f}_1, \mathbf{f}_2, \cdots, \mathbf{f}_n\}$, these procedures can be noted as:
\begin{equation}
\begin{aligned}
    \mathbf{f}^{\prime}_i & = \mathcal{F}^{trans}_i(\mathbf{f}_i)|_{i=1}^n,~\text{where}~\mathbf{f}^{\prime}_i \in \mathrm{R}^d, \\
    \boldsymbol \alpha & = \mathcal{F}^{gate}(\text{Concat}(\mathbf{f}_1, \mathbf{f}_2, \cdots, \mathbf{f}_n)),~\text{where}~ \boldsymbol \alpha \in \mathrm{R}^{n}, \\
    \mathbf{g} & = \sum\nolimits_{i=1}^n \mathbf{f}^{\prime}_i \cdot \alpha_i, ~\text{where}~\mathbf{g} \in \mathrm{R}^d,
\end{aligned}
\end{equation}
where $\mathcal{F}^{trans}_i(\cdot)$ denotes the mapping function for the $i$-th transformation block, and $\mathcal{F}^{gate}(\cdot)$ represents the mapping function for the gating network. And $d$ is the number of aligned feature dimensions. 

In addition, to emphasize the importance of critical features and enhance the generalization ability, we propose to impose a sparsity constraint as a regularization on the distribution of gating weights. The $\mathcal{L}_1$ loss is utilized to penalize non-zero weights resulting in more weights near zero. This constraint can be written as:
\begin{equation}
    \text{min}~\mathcal{L}_{sparse}(\boldsymbol \alpha) = \text{min}~\sum\nolimits_{i=1}^n ||\alpha_i||_1.
\end{equation}
In this way, QAM allows for capturing a broader range of quality-related features, thereby enabling better adaptation to various types of video content, distortions, or movement. Then the aggregated feature is sent into a regression head, which is a single fully-connected layer, for quality prediction, resulting in $y^t$.

\subsection{Optimization Objective}\label{sec:objective}

In practical scenarios, using these large pretrained models for inference can be computationally expensive. To reduce the computational cost and increase flexibility, we propose to use knowledge distillation \cite{DBLP:journals/corr/HintonVD15/ori-kd} to transfer the knowledge from large and complex models to a lightweight VQA model. 

In this paper, a Video Swin Transformer-Tiny \cite{DBLP:journals/corr/abs-2106-13230/video-swin} is selected as the backbone. For an input video $\mathcal{V}$, the quality representation can be achieved by $\mathbf{h}=\mathcal{H}(\mathcal{V})$, where $\mathcal{H}(\cdot)$ represents the mapping function of the lightweight backbone, and $\mathbf{h}\in \mathrm{R}^{d}$. Then $\mathbf{h}$ is sent into a regression head, which is a single fully-connected layer, for quality prediction, resulting in $y^s$. Note that both $y^t$ and $y^s$ are supervised by the labeled MOS using a smooth $\mathcal{L}_1$ loss. Additionally, we apply a similarity consistency in knowledge distillation between $\mathbf{g}$ and $\mathbf{h}$. This allows the VQA model to simulate the robust quality representation generated by diverse pretrained models, further enhancing its performance. Given the labeled MOS $y$, the regression loss for pretrained models can be noted as:
\begin{equation}
    \mathcal{L}_{reg}^t(y,y^t) = \left \{
        \begin{array}{cc}
             0.5(y-y^t)^2, & \text{if}~|y-y^t|<1 \\
             |y-y^t|-0.5,  & \text{otherwise.}
        \end{array}
    \right.
\end{equation}
And the regression loss for the lightweight VQA model $\mathcal{L}_{reg}^s(y,y^s)$ share the same formulation. A $\mathcal{L}_2$ loss is used for the process of knowledge distillation, which can be written as:
\begin{equation}
    \mathcal{L}_{kd}(\mathbf{g},\mathbf{h}) = ||\mathbf{g}-\mathbf{h}||_2.
\end{equation}
The whole optimization objective can be formulated as:
\begin{equation}\label{equ:optimization}
    \text{min}~\mathcal{L}_{reg}^s(y,y^s) + \gamma\big(\mathcal{L}_{reg}^t(y,y^t)+\mathcal{L}_{kd}(\mathbf{g},\mathbf{h})\big) + \lambda \mathcal{L}_{sparse}(\boldsymbol \alpha),
\end{equation}
where $\gamma$ is a balancing weight for knowledge distillation, and $\lambda$ is a hyper-parameter to balance the level of sparsity.

\section{Experiments}

\subsection{Dataset and Evaluation Metrics}

\paragraph{Dataset.} Our method is evaluated on three widely-adopted public NR-VQA datasets, including KoNViD-1k \cite{DBLP:conf/qomex/HosuHJLMSLS17/konvid}, LIVE-VQC \cite{DBLP:journals/tip/SinnoB19/live-vqc}, and YouTube-UGC \cite{DBLP:conf/mmsp/WangIA19/youtube-ugc}. Mean opinion scores (MOS) are provided along with training videos. 
Specifically, KoNViD-1k contains 1,200 videos that are fairly filtered from a large public video dataset YFCC-100M. The time duration of the video is 8 seconds. And these videos have a frame rate of 24/25/30 FPS and a resolution of $960\times 540$.
LIVE-VQC consists of 585 videos with complex authentic distortions, which are captured by 80 users using 101 different devices, ranging from 240P to 1080P.
YouTube-UGC has 1,380 UGC videos sampled from YouTube with a duration of 20 seconds and resolutions from 360P to 4K.
All these datasets contain no pristine videos, thus only NR methods can be evaluated on them.
Following \cite{DBLP:conf/mm/XuLZZW021/stdm}, we split all the dataset into 80\% training videos and 20\% testing videos randomly.

\paragraph{Evaluation Metric.} Spearman’s Rank-Order Correlation Coefficient (SRCC) and Pearson’s Linear Correlation Coefficient (PLCC) are selected as metrics to measure the monotonicity and accuracy, respectively. They are in the range of 0.0 to 1.0, and larger values indicate better results.  Besides, the mean average of PLCC and SRCC is also reported as a comprehensive criterion.

\subsection{Implementation Details}

Our method is implemented based on PyTorch \cite{DBLP:conf/nips/PaszkeGMLBCKLGA19} and MMAction2 \cite{2020mmaction2}. All experiments are conducted on 4 NVIDIA V100 GPUs.
For all datasets, we select EfficientNet-b7, ir-CSN-152, CLIP,  Swin Transformer Base, TimeSformer, Video Swin Transformer Base and SlowFast as candidate pretrained models, and choose the Video Swin Transformer Tiny as the lightweight VQA model. 
Frames are sampled in each video with a fixed temporal step to form a clip input. For frame-wise models (\eg EfficientNet, CLIP), the feature representation can be calculated through the average features of all frames. For video clip-based models (\eg SlowFast, ir-CSN-152), the extracted features can be used directly for the video representation.
For KoNViD-1k, we sample 16 frames with a frame interval of 2.
As videos in LIVE-VQC and YouTube-UGC have longer time durations, we sample 64 frames with an interval of 2, and 32 frames with an interval of 8, respectively. 
Since most augmentations will introduce extra interference to the quality of videos (\eg resize, color jitter) \cite{DBLP:journals/corr/abs-2108-05997/musiq}, we only choose the center crop to produce inputs with a resolution of $224 \times 224$. 
During the optimization procedure, we use the AdamW optimizer with a weight decay of 2e-2. 
A cosine annealing scheduler with a warmup of 2 epochs is adopted to control the learning rate. The initial learning rate is 1e-3. And $\gamma$ is 0.1 by default. $\lambda$ is set to 0.8. $d$ is set to 32. The batch size of the input is set to 1. All models are trained for 60 epochs. And the checkpoint generated by the last iteration is used for evaluation. 
For inference, we follow a similar procedure as \cite{DBLP:journals/corr/abs-2103-15691/vivit} by using $4\times 5$ views. In the procedure, a video is uniformly sampled as 4 clips in the temporal dimension, and for each clip, the shorter spatial side is scaled to 256 pixels and we take 5 crops in the four corners and the center. The final score is computed as the average score of all the views. The average result of 10 repeat runs with different random splits is used as the final score for the experiments in Tab.~\ref{tab:sota}.

\begin{table*}[t]
\caption{
    Comparisons with SOTA methods. The up arrow ``$\uparrow$" means that a larger value indicates better performance. The mark ``-'' means the results are not reported originally. Mark ``*" indicates that the model uses external QA data for training. The best and second best performances are \textbf{highlighted} and \underline{underlined}. Ada-DQA outperforms existing SOTA methods by large margins on three datasets. We also report the performance of the Video Swin Tiny without the assistance of diverse pretrained models.
}
\label{tab:sota}
\centering
\begin{tabular}{c|ccc|ccc|ccc}
\toprule
\multirow{2}{*}{Method} & \multicolumn{3}{c|}{KoNViD-1k} & \multicolumn{3}{c|}{LIVE-VQC} & \multicolumn{3}{c}{YouTube-UGC} \\
& SRCC$\uparrow$ & PLCC$\uparrow$ & Mean
& SRCC$\uparrow$ & PLCC$\uparrow$ & Mean
& SRCC$\uparrow$ & PLCC$\uparrow$ & Mean \\
\midrule
VIIDEO \cite{DBLP:journals/tip/MittalSB16/viideo} 
& 0.2980 & 0.3030 & 0.3005 
& 0.0332 & 0.2164 & 0.1248 
& 0.0580 & 0.1534 & 0.1057 \\
NIQE \cite{DBLP:journals/spl/MittalSB13/niqe} 
& 0.5417 & 0.5530 & 0.5474
& 0.5957 & 0.6286 & 0.6122
& 0.2379 & 0.2776 & 0.2578 \\
BRISQUE \cite{DBLP:journals/tip/MittalMB12/spatial-domain} 
& 0.654 & 0.626 & 0.640
& 0.592 & 0.638 & 0.615
& 0.382 & 0.395 & 0.389\\
VSFA \cite{DBLP:conf/mm/LiJJ19/vsfa} 
& 0.755 & 0.744 & 0.750
& - & - & -
& - & - & - \\
TLVQM \cite{DBLP:conf/mmsp/EbenezerSWWB20/tlvqm} 
& 0.7729 & 0.7688 & 0.7709
& 0.7988 & 0.8025 & 0.8807
& 0.6693 & 0.6590 & 0.6642 \\
RIRNet \cite{DBLP:conf/mm/ChenLMWS20/rirnet} 
& 0.7755 & 0.7812 & 0.7784
& 0.7713 & 0.7982 & 0.7848
& - & - & - \\
MDTVSFA \cite{DBLP:journals/ijcv/LiJJ21} 
& 0.7812 & 0.7856 & 0.7834
& 0.7382 & 0.7728 & 0.7555
& - & - & - \\
RIRNet+\emph{CSPT} \cite{DBLP:journals/tip/ChenLWDS22/cspt} 
& 0.8008 & 0.8062 & 0.8035
& 0.7989 & 0.8194 & 0.8092
& - & - & -\\
CoINVQ \cite{DBLP:conf/cvpr/WangKTYBAMY21/rich-feats} 
& 0.802 & 0.816 & 0.809
& - & - & -
& 0.764 & 0.767 & 0.766 \\
RAPIQUE \cite{DBLP:journals/corr/abs-2101-10955/rapique} 
& 0.8031 & 0.8175 & 0.8103
& 0.7548 & 0.7863 & 0.7706
& 0.7591 & 0.7684 & 0.7638 \\
StarVQA \cite{DBLP:journals/corr/abs-2108-09635/star-vqa} 
& 0.812 & 0.796 & 0.804
& 0.732 & 0.808 & 0.770
& - & - & - \\
BVQA* \cite{DBLP:journals/corr/abs-2108-08505/bvqa} 
& 0.8362 & 0.8335 & 0.8349
& \underline{0.8412} & \underline{0.8415} & \underline{0.8414}
& 0.8312 & 0.8194 & 0.8253 \\
STDAM \cite{DBLP:conf/mm/XuLZZW021/stdm} 
& 0.8448 & 0.8415 & 0.8432
& 0.7931 & 0.8204 & 0.8068
& \underline{0.8341} & \underline{0.8297} & \underline{0.8319} \\
DisCoVQA \cite{DBLP:journals/corr/abs-2206-09853}
& 0.847 & 0.847 & 0.847
& 0.820 & 0.826 & 0.823
& - & - & - \\
FastVQA \cite{DBLP:conf/eccv/WuCHLWSYL22} 
& \underline{0.859} & \underline{0.855} & \underline{0.857}
& 0.823 & 0.844 & 0.834
& - & - & -\\
\midrule
Video Swin Tiny
& 0.8316 & 0.8694 & 0.8505
& 0.8335 & 0.8316 & 0.8326
& 0.8566 & 0.8499 & 0.8533 \\
Ada-DQA 
& \textbf{0.8651} & \textbf{0.8831} & \textbf{0.8741}
& \textbf{0.8591} & \textbf{0.8587} & \textbf{0.8589}
& \textbf{0.8729} & \textbf{0.8800} & \textbf{0.8765} \\
\midrule
$\Delta$ than the previous best 
& +0.6\% & +2.8\% & +1.7\% 
& +1.79\% & +1.72\% & +1.75\%
& +3.88\% & +5.03\% & +4.46\%
\\
\midrule
$\Delta$ than \textit{w/o} pretrained models 
& +3.35\% & +1.37\% & +2.36\% 
& +2.56\% & +2.71\% & +2.63\%
& +1.63\% & +3.01\% & +2.32\%
\\
\bottomrule
\end{tabular}
\end{table*}

\subsection{Comparison with SOTA methods}

We report the SRCC and PLCC performance with current SOTA methods on KoNViD-1k, LIVE-VQC, and YouTube-UGC. 
As shown in Tab.~\ref{tab:sota}, \textbf{our method achieves new state-of-the-art results on all these datasets without using extra training data of QA}.
Some observations can also be found through these results.
We can observe those deep learning-based methods outperform the traditional hand-crafted method (\eg VIIDEO, NIQE) largely.
Besides, within deep learning-based methods, VQA methods produce much better performance than IQA methods (\eg BRISQUE). Since there exist many temporal-distributed distortions in these datasets, VQA models can capture this temporal information.
Ada-DQA outperforms StarVQA, which builds a Transformer model to capture spatiotemporal information, in large margins.
BVQA incorporates extra training data from other QA datasets for the feature extractor. And our method still \textbf{outperforms it without using external training data of QA} (+2.89\% of SRCC, and + 4.96\% of PLCC in KoNViD-1k). This shows the advantage of leveraging quality-related knowledge from pretrained models.
Compare with the current best method of STDAM, DiscoVQA, and FastVQA, Ada-DQA improves the best performance in \textbf{large margins} to $0.8651\pm 0.0034$ of SRCC in KoNVid-1k (+0.6\%), $0.8591\pm 0.0041$ of SRCC in LIVE-VQC (+1.79\%) and $0.8729\pm 0.0086$ of SRCC in YouTube-UGC (+3.88\%). The accuracy and consistency of prediction results have been significantly improved.

\subsection{Experimental Analysis and Ablation Studies}

\begin{table}[t]
    \centering
    \caption{
        Experimental analysis on different numbers of selected pretrained models \textit{with} or \textit{without} the usage of sparsity constraint in QAM. SRCC results are reported. The best result under the setting of \textit{with} or \textit{without} is bolded.
    }
    \begin{tabular}{cc|ccc}
    \toprule
        Number & QAM & KoNViD-1k & LIVE-VQC & YouTube-UGC \\
    \midrule
        3 & $\times$ & 0.8517 & 0.8432 & 0.8630 \\
        4 & $\times$ & 0.8613 & \textbf{0.8510} & \textbf{0.8659} \\
        5 & $\times$ & 0.8601 & 0.8490 & 0.8591 \\
        6 & $\times$ & \textbf{0.8615} & 0.8448 & 0.8589 \\
        7 & $\times$ & 0.8573 & 0.8459 & 0.8621 \\
    \midrule
        3 & \checkmark & 0.8432 & 0.8433 & 0.8621 \\
        4 & \checkmark & 0.8623 & 0.8514 & 0.8695 \\
        5 & \checkmark & 0.8621 & 0.8546 & 0.8679 \\
        6 & \checkmark & 0.8645 & 0.8542 & 0.8644 \\
        7 & \checkmark & \textbf{0.8651} & \textbf{0.8591} & \textbf{0.8729} \\
    \midrule
        8 & \checkmark & 0.8641 & 0.8560 & 0.8711 \\
    \bottomrule
    \end{tabular}
    \label{tab:number}
\end{table}

\begin{table}[t]
    \centering
    \caption{
        Experiments using a single pretrained model for knowledge distillation. The SRCC results in three datasets are reported. The best results in different datasets are bolded.
    }
    \begin{tabular}{c|ccc}
    \toprule
        Pretrained Model & {KoNViD-1k} & {LIVE-VQC} & {YouTube-UGC} \\
    \midrule
        \textit{w/o} & 0.8316 & 0.8335 & 0.8566 \\
    \midrule
        EfficientNet-b7 & {0.8412} & {0.8495} & 0.8587 \\
        Video Swin Base & 0.8390 & 0.8173 & {0.8603} \\
        Swin Base & 0.8391 & {0.8475} & 0.8497 \\
        TimeSformer & 0.8409 & 0.8302 & \textbf{0.8618} \\
        CLIP & 0.8404 & 0.8341 & {0.8608} \\
        ir-CNS-152 & \textbf{0.8458} & \textbf{0.8521} & 0.8518 \\
        SlowFast & {0.8423} & 0.8041 & 0.8523 \\  
    \bottomrule
    \end{tabular}
    \label{tab:single}
\end{table}

\paragraph{Number of pretrained models and effectiveness of sparsity constraint in QAM} To investigate the impact of the number of pretrained models, we performed experiments by reducing the number of models from 7 to 3. As depicted in Tab.~\ref{tab:number}, increasing the number of models does not always lead to better performance without the use of sparsity constraints in QAM. \textbf{With the help of sparsity constraint, the model can achieve continuous improvement as the number of pretrained models increases}. However, adding more models beyond 8 (introducing an extra model of ViT Base) does not yield any further improvements. This may indicate that the quality-related information provided by the pretrained models has reached a saturation point. Therefore, we set the number of pretrained models to 7 in our experiments. In these experiments, the pretrained model is randomly selected and removed.

\paragraph{The necessary of diverse pretrained models} Moreover, we also conduct experiments solely utilizing a singular pretrained model to demonstrate the disparity among models. As depicted in Tab.~\ref{tab:single}, a solitary pretrained model cannot consistently attain optimal outcomes across all VQA datasets. For instance, while CLIP excels on YouTube-UGC, it falls considerably short on the other two datasets. We posit that this is influenced by the correlation between pretrained models, such as pre-text task, dataset, architecture, and VQA tasks. Additionally, the results obtained by utilizing singular pretrained models are notably distant from the state-of-the-art. These findings substantiate that a solitary pretrained model is inadequate for diverse application scenarios, and \textbf{leveraging a variety of pretrained models is imperative}.

\begin{table}[t]
    \centering
    \caption{Experiments on different distillation losses.}
    \label{tab:ab_kdloss}
    \begin{tabular}{c|cc}
    \toprule
        distillation loss & SRCC$\uparrow$ & PLCC$\uparrow$ \\
    \midrule
        $\mathcal{L}_2$ & \textbf{0.8651} & \textbf{0.8831} \\
    \midrule
        $\mathcal{L}_1$ & 0.8092 & 0.8310 \\
        Jesen-Shannon & 0.8455 & 0.8646 \\
    \bottomrule
    \end{tabular}
\end{table}

\paragraph{Different types of distillation loss} Experiments on different knowledge distillation losses are performed in KoNViD-1k, including the $\mathcal{L}_2$ loss, the $\mathcal{L}_1$ loss and the Jesen-Shannon (JS) loss \cite{DBLP:conf/isit/FugledeT04/js-divergence}. As shown in Tab.~\ref{tab:ab_kdloss}, the $\mathcal{L}_2$ loss owns the best performance in transferring aggregated features by multiple teacher models.

\paragraph{Selection of hyper-parameters} We conduct experiments to show how the hyper-parameters $\gamma$ and $\lambda$ in Equ.~\ref{equ:optimization} will influence the final results in KoNViD-1k. The results are listed in Tab.~\ref{tab:gamma} and \ref{tab:lambda}. When $\gamma$ is 0.1, and $\lambda$ is 0.8, which are used in our experiments, the best performance can be obtained.

\begin{table}[t]
    \caption{Selection of the hyper-parameters of $\gamma$ and $\lambda$.}
    \label{tab:hyperparam}
    \begin{subtable}[t]{0.23\textwidth}
        \centering
        \caption{hyper-parameter $\gamma$}
        \label{tab:gamma}
        \begin{tabular}{c|cc}
            \toprule
                $\gamma$ & SRCC$\uparrow$ & PLCC$\uparrow$ \\
            \midrule
                0.1 & \textbf{0.8651} & \textbf{0.8831} \\
            \midrule
                0.2 & 0.8462 & 0.8663 \\
                0.5 & 0.8406 & 0.8698 \\
                1.0 & 0.8320 & 0.8549 \\
            \bottomrule
        \end{tabular}
    \end{subtable}
    \hfill
    \begin{subtable}[t]{0.23\textwidth}
        \centering
        \caption{hyper-parameter $\lambda$.}
        \label{tab:lambda}
        \begin{tabular}{c|cc}
            \toprule
                $\lambda$ & SRCC$\uparrow$ & PLCC$\uparrow$ \\
            \midrule
                0.8 & \textbf{0.8651} & \textbf{0.8831} \\
            \midrule
                0.2 & 0.8449 & 0.8671 \\
                0.5 & 0.8528 & 0.8716 \\
                1.0 & 0.8602 & 0.8780 \\
            \bottomrule
        \end{tabular}
    \end{subtable}
\end{table}


\begin{table}[t]
    \centering
    \caption{Contributions of different pretrained models by $\alpha$.}
    \label{tab:alpha}
    \begin{tabular}{c|cc}
    \toprule
        Model & $\alpha$ for LQ videos & $\alpha$ for HQ videos \\
    \midrule
        EfficientNet-b7 & 0.1303 & 0.7208 \\
        Video Swin Base & 0.2021 & 0.0455 \\
        Swin Base       & 0.1676 & 0.6805 \\
        TimeSformer     & 0.4679 & 0.2317 \\
        CLIP            & 0.1279 & 0.3567 \\
        ir-CNS-152      & 0.8690 & 0.2788 \\
        SlowFast        & 0.4210 & 0.0018 \\  
    \bottomrule
    \end{tabular}
\end{table}

\paragraph{Contribution of different pretrained models} To investigate the contribution, we analyze the gating weights $\boldsymbol{\alpha}$ generated by the QAM in KoNViD-1k. The statistical average scores for different models are calculated. 
We count responses of low-quality (LQ, MOS<3.5) and high-quality (HQ, MOS>3.5) videos in Tab.~\ref{tab:alpha}. It can be seen that \textbf{for LQ videos, models that can provide distortion and motion-related information (\eg ir-CSN-152) have larger weights; for HQ videos, models that can provide content-related one (\eg EfficientNet-b7) own larger weights}.

\paragraph{Computational cost} We compare the \#Params, \#FLOPs, and SRCC of Ada-DQA over some SOTA methods whose models are available: Ada-DQA (29M, 88T, 0.8651), MDTVSFA (24M, 168T, 0.7812), StarVQA (121M, 75T, 0.812), BVQA (24M, 240T, 0.8362). With the help of pretrained models during training, Ada-DQA obtains higher results with a fair cost during inference.


\subsection{Visualization of the Attention}

\begin{figure}[t]
\begin{center}
   \includegraphics[width=0.8\linewidth]{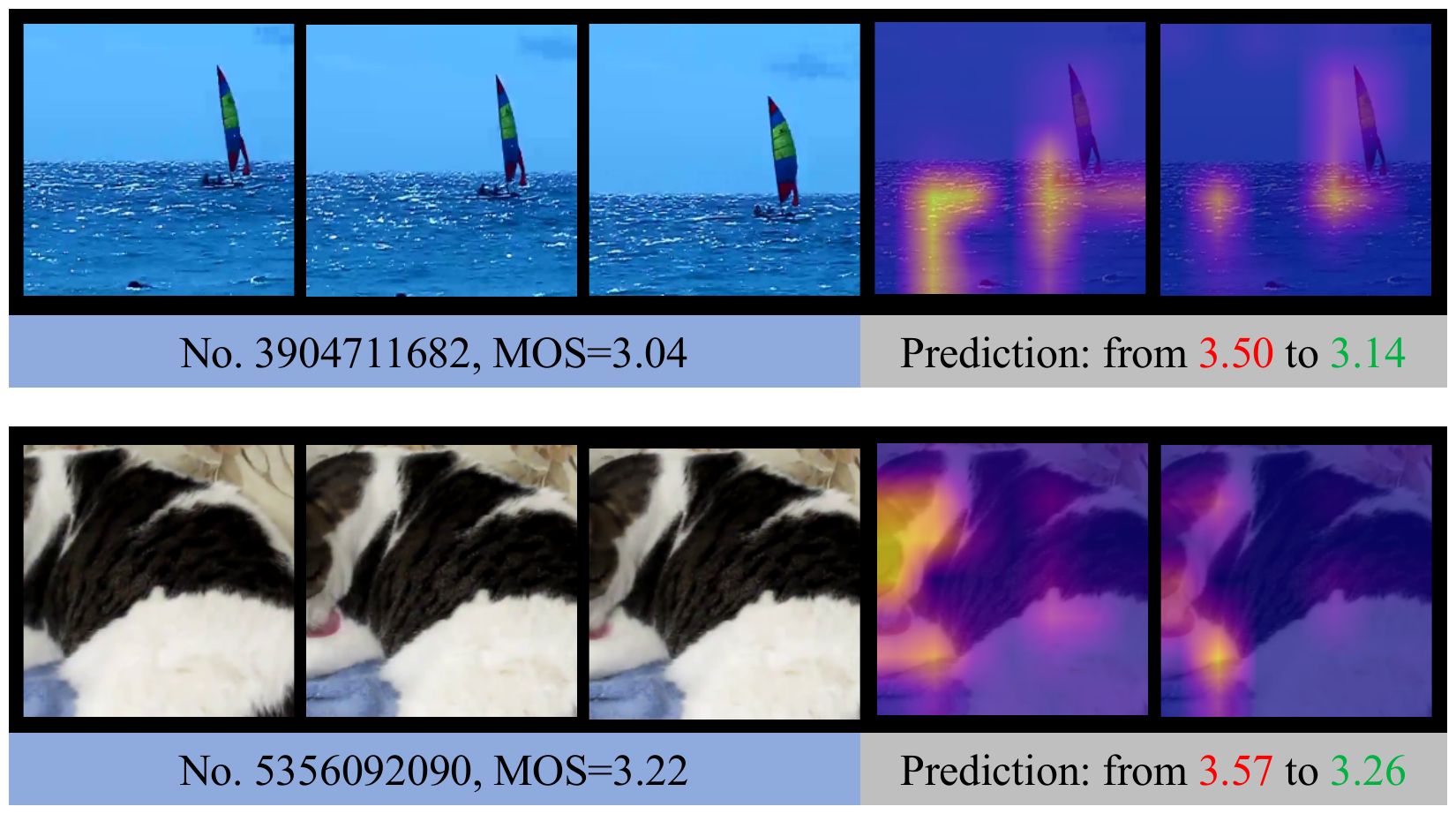}
\end{center}
\caption{
Comparison of predictions and attention visualizations. For each video, video frames (left) and attention maps (right) \textcolor{red}{before} and \textcolor{green}{after} using Ada-DQA are illustrated.}
\label{fig:visualization}
\end{figure}

Some representative videos in KoNViD-1k are selected to show the performance improvement brought by Ada-DQA. Visualization of the feature attention maps using GradCAM \cite{DBLP:conf/iccv/SelvarajuCDVPB17/grad-cam} are shown in Fig.~\ref{fig:visualization}. After introducing the adaptive acquisition strategy, Ada-DQA generates more accurate results and the attention maps highlight more quality-related regions.
For instance, in the first video clip, attention from the vast surface of the ocean (left) is transferred to the sailboat, with some attention on ocean waves kept (right). Ada-DQA focuses on areas more related to the action (boat sailing) or giving clues about the perceptual quality (edges of waves). 

\section{Conclusion}

To address the issue of insufficient training data in VQA, this paper analyzes the entire spectrum of video distribution diversity that impacts quality and proposes the Ada-DQA framework, which employs a range of diverse pretrained models to improve quality representation. With Ada-DQA, it becomes possible to extract critical and relevant features generated by different frozen pretrained models adaptively. Experimental results on three mainstream NR-VQA benchmarks show the effectiveness in the context of limited data. Thorough analysis and ablation studies also validate the necessity of each component. This work hopes to inspire future research that leverages pretrained models to aid in a wider array of tasks.

\section*{Acknowledgments}

This research was partly supported by the National Key R\&D Program of China (Grant No. 2020AAA0108303), and Shenzhen Science and Technology Project (Grant No. JCYJ20200109143041798) and Shenzhen Stable Supporting Program (No. WDZC20200820200655001) and Shenzhen Key Laboratory of next-generation interactive media innovative technology (No. ZDSYS20210623092001004).

\bibliographystyle{ACM-Reference-Format}
\bibliography{paper}










\end{document}